\documentclass[11pt]{article}
\usepackage[margin=1in]{geometry} 
\usepackage{amsmath,amsfonts}
\usepackage{algorithmic}
\usepackage{array}
\usepackage[caption=false,font=normalsize,labelfont=sf,textfont=sf]{subfig}
\usepackage{textcomp}
\usepackage{stfloats}
\usepackage{url}
\usepackage{verbatim}
\usepackage{graphicx}
\usepackage{tabularx}
\usepackage{balance}

\usepackage{cite}
\usepackage{bm}
\usepackage{amsthm}
\usepackage{amssymb}
\usepackage{dsfont}
\usepackage{algorithm}
\usepackage{wrapfig}
\usepackage{multirow}
\usepackage{threeparttable}
\usepackage{lipsum}
\usepackage{hyperref}
\usepackage{bbding}

\usepackage{caption}
\newcommand{\tbeta}{\widetilde{\boldsymbol{\beta}}}
\newcommand{\hfv}{\widehat{f}_{v}}
\newcommand{\hf}{\widehat{f}}

\newcommand{\bbeta}{\bm{\beta}}
\newcommand{\bhbeta}{\widehat{\bbeta}}

\newcommand{\ind}[1]{\mathds{I}[#1]}

\newcommand{\Bc}{\mathcal{B}}

\newcommand{\D}{\mathcal{D}}

\newcommand{\Rb}{\mathbb{R}}
\newcommand{\e}{\mathbf{e}}

\newcommand{\bI}{\mathbf{I}}

\newcommand{\Mcc}{\mathcal{M}}

\newcommand{\R}{\mathbf{R}}

\newcommand{\Sc}{\mathcal{S}}

\renewcommand{\u}{\mathbf{u}}

\newcommand{\X}{\mathbf{X}}

\newcommand{\x}{\mathbf{x}}

\newcommand{\y}{\mathbf{y}}

\newcommand{\abs}[1]{\mbox{$\lvert #1 \rvert$}}

\newtheorem{thm}{Theorem}

\newtheorem{defn}{Definition}

\newtheorem{rem}{Remark}

\newtheorem{assum}{Assumption}

\newcommand{\vbeta}{\boldsymbol \beta}

\newcommand{\hvbeta}{\widehat{\boldsymbol{\beta}}}

\title{Efficient Sparse Least Absolute Deviation Regression with Differential Privacy}
\author{Weidong Liu\thanks{School of Mathematical Sciences, MoE Key Lab of Artificial Intelligence, Shanghai Jiao Tong University, Shanghai, 200240, China (e-mail: weidongl@sjtu.edu.cn).} , 
        Xiaojun Mao\thanks{School of Mathematical Sciences, Ministry of Education Key Laboratory of Scientific and Engineering Computing, Shanghai Jiao Tong University, Shanghai, 200240, China (e-mail: maoxj@sjtu.edu.cn).} ,
        Xiaofei Zhang\thanks{School of Statistics and Mathematics, Zhongnan University of Economics and Law, Wuhan, 430073, China (e-mail: zhangxf@zuel.edu.cn).} , 
       and Xin Zhang\thanks{Department of Statistics, Iowa State University, Ames, 50011, USA (e-mail: xinzhang.nac@gmail.com).}
}

\date{}

\begin{document}

\maketitle

\begin{abstract}
In recent years, privacy-preserving machine learning algorithms have attracted increasing attention because of their important applications in many scientific fields. However, in the literature, most privacy-preserving algorithms demand learning objectives to be strongly convex and Lipschitz smooth, which thus cannot cover a wide class of robust loss functions (e.g., quantile/least absolute loss). In this work, we aim to develop a fast privacy-preserving learning solution for a sparse robust regression problem. Our learning loss consists of a robust least absolute loss and an $\ell_1$ sparse penalty term. To fast solve the non-smooth loss under a given privacy budget, we develop a Fast Robust And Privacy-Preserving Estimation (FRAPPE) algorithm for least absolute deviation regression. Our algorithm achieves a fast estimation by reformulating the sparse LAD problem as a penalized least square estimation problem and adopts a three-stage noise injection to guarantee the $(\epsilon,\delta)$-differential privacy. We show that our algorithm can achieve better privacy and statistical accuracy trade-off compared with the state-of-the-art privacy-preserving regression algorithms. In the end, we conduct experiments to verify the efficiency of our proposed FRAPPE algorithm.
\end{abstract}


\section{Introduction}

Recent years have witnessed that machine learning (ML) technology has rapidly and fundamentally changed human daily lives. However, as many ML models and training procedures involve sensitive personal information (e.g. healthcare data, financial records, web browser history, etc.), data privacy becomes a crucial concern.
To ensure that only population-wise patterns are revealed, but individual sensitive information is well protected, differential privacy has been established \cite{dwork2014algorithmic} as a rigorous and standard definition of privacy.
Under the framework of differential privacy, a long line of research on privacy-preserving ML has been inspired, such as sparse regression \cite{wang2019sparse,cai2021cost,wang2018high}, clustering analysis \cite{ren2017dplk,mulle2015privacy}, and deep neural network training \cite{abadi2016deep,ha2019differential,zhu2020private}.

Although various differential private ML methods have been developed, few of them take the model or algorithm robustness into consideration.
In real-world ML applications, data are typically contaminated with heavy-tailed noises and abnormal values \cite{Horowitz98}.
Naively analyzing the data with classical least square loss will lead to a fitting bias and deteriorate the estimation performance.
To mitigate the negative impact of outlier data and improve learning robustness, in the literature, the least absolute loss has become a popular surrogate for the least square loss in regression analysis. 
As the sparse regression problem covers a broad class of supervised ML tasks, there is a pressing demand to develop an efficient privacy-preserving algorithm for sparse least absolute deviation (LAD) regression.

Unfortunately, the design of an efficient privacy-preserving sparse LAD regression algorithm is a non-trivial task.
On the one hand, to improve the privacy of an iterative learning algorithm, a widely adopted strategy is to inject certain unbiased independent random noise into the intermediate statistics/variables during iterations \cite{jayaraman2018distributed,zhang2017efficient,wang2019efficient, zhang2020private}.
However, such a noise injection method typically deteriorates the privacy guarantee as the iteration round $T$ increases.  
Therefore, with a given privacy budget, there is an upper limit for the total iteration rounds, which introduces the trade-off between privacy and accuracy.
On the other hand, unlike the classical least square loss, the least absolute loss is a non-smooth function with no Lipschitz continuous gradient.
Thus, the commonly used Newton or gradient-descent based algorithms are not applicable to the LAD regression problem.
To solve the non-smooth least absolute loss, a few algorithms were proposed in recent literature, including the subgradient descent algorithm \cite{wang2017distributed,wang2019distributed}, 
EM algorithm\cite{tian2014linear,tian2016estimation,zhou2014quantile}, ADMM\cite{Yu-Lin17,Gu-Fan-Kong18} and smoothed approximations \cite{He-Pan-Tan21} 
But these algorithms suffer poor convergence performance, which means a large value of iterations $T$ is needed for the desired estimation accuracy.
The limitations caused by these two conflicting challenges naturally motivate our research question: 
\emph{Is it possible to develop an efficient privacy-preserving algorithm for sparse LAD regression, so that it could achieve excellent performance in terms of both privacy protection and estimation accuracy?}

In this paper, we will give an \emph{affirmative} answer to the above question. 
We develop a novel algorithm termed FRAPPE for fast robust and privacy-preserving sparse LAD regression estimation.
Our main contributions and their significance are summarized in the following:
\begin{itemize}
    \item We study the robust sparse regression problem with privacy constraints.
    Our learning loss is a combination of non-smooth least absolute loss and $\ell_1$ sparse penalty.
    To optimize the loss efficiently, we propose an efficient algorithm, named FRAPPE. 
    Specifically, our algorithm consists of three steps: initial estimation, pseudo response transformation, and surrogate loss optimization. 
    Among them, the key step is the pseudo response transformation. 
    Our proposed transformation converts the least absolute loss to the least square loss.
    Therefore, the problem can be easily solved with a proximal gradient descent based algorithm.
    
    \item To guarantee $(\epsilon,\delta)$-differential privacy, our algorithm is designed with a three-stage noise injection: 
    The first stage of noise injection is for the initial estimator. 
    The output perturbation is applied so that the procedure does not depend on the initialization method.
    The second stage is to add random noise to the kernel density estimation, which is essential for constructing pseudo responses.
    In the third stage, gradient perturbation is adopted as our algorithm leverages the proximal gradient descent for surrogate loss optimization. 

    \item By incorporating the convergence error and injected noises, we establish the 
    statistical consistency rate with the privacy-accuracy trade-off in Theorem \ref{thm:accuracyv}. 
    We show in Theorem \ref{thm:accuracyv} that after a constant number of iterations, our estimators achieve a trade-off between the near-optimal rate of $O(\sqrt{s\log p/N})$ and the privacy loss $O(\sqrt{p\log(1/\delta)\log(N\epsilon)}/(N \epsilon))$, where $N, p, s$ are the size of the total sample, dimension of regression weights, and number of non-zero weights.
\end{itemize}
The rest of the paper is organized as follows. Section \ref{sec:related_work} discusses related works for robust regression estimation and differential private sparse regression. In Section \ref{sec:algorithm}, we specify the problem formulation and introduce the proposed FRAPPE algorithm. Section \ref{sec:theory} shows the privacy guarantee of the proposed algorithm and the theoretical guarantee of statistical consistency and sparsity recovery. Simulation results and real data analysis are reported in Section \ref{sec:simulation}. The conclusion is given in Section \ref{sec:conclusion}, and the technical details are given in the supplementary materials.

\textbf{Notation:} For any $p$ dimensional vector $\x = (x_1, \cdots, x_p)^\top$, we use $\|\x\|_1 = \sum_{i=1}^p \abs{x_i}$ to denote the $\ell_{1}$-norm, and $\|\x\|_2 = (\sum_{i=1}^p x_i^2)^{1/2}$ to denote the $\ell_{2}$-norm. Given two sequences $a_n$ and $b_n$, we define $a_n = O(b_n)$ if there is a constant $M > 0$ such that $a_n \leq M b_n$. We say $a_{n} = \Theta(b_{n})$ if and only if $a_n = O(b_n)$ and $b_n = O(a_n)$ both hold. Given a sequence of random variables $X_n$ and a sequence of constants $a_n$, we say $X_n = O_{\mathbb{P}}(a_n)$ if for any $\varepsilon > 0$, there exist a finite $M > 0$ and a finite $N > 0$ such that $P\left(\left|X_{n} / a_{n}\right|>M\right)<\varepsilon, \forall n>N$.

\section{Preliminaries and Related Works} \label{sec:related_work}

\subsection{Robust Linear Regression}

The linear regression model has been one of the most powerful machine learning tools. Given the independently and identically distributed observations $\{\x_i,y_i\}_{i=1}^{N}$, we assume that the data follow $y_i= \x_i^\top \vbeta^* + e_i$, where $\vbeta^*$ is an unknown regression weight and $e_i$ is the corresponding random error for the $i$th observation.
Traditionally, the regression weight $\vbeta^*$ can be estimated by solving the ordinary least squares (OLS) estimation problem $\sum_{i=1}^{N}(y_i-\x_i^\top\vbeta)^2/N$.
However, it is commonly known that the OLS estimation is very sensitive to outlier data, and thus it is not a robust estimator \cite{huber2011robust,yu2017robust}. 

In the literature, various methods have been proposed to enhance the robustness of regression, including M-estimates \cite{huber2011robust}, least median of squares (LMS) estimates \cite{siegel1982robust}, least trimmed squares (LTS) estimates \cite{rousseeuw1993alternatives} and S-estimates \cite{rousseeuw1984robust}, etc.
In this work, we focus on the least absolute deviation (LAD) regression method to achieve robust regression estimation.
LAD regression method, also known as the median regression method, estimates the coefficient $\vbeta^*$ by minimizing the least absolute loss $\sum_{i=1}^{N}|y_i-\x_i^\top\vbeta|/N$.
Note that the LAD regression can be viewed as a special case of the well-known quantile regression by taking the quantile level as $1/2$.
Quantile regression has been widely adopted for robust estimation as its estimator is less likely affected by the outliers than OLS estimator \cite{Li-Zhu08,Belloni-Chernozhukov11,Wang-Wu-Li12,Zheng-Peng-He18}.

Furthermore, we also consider the problem of identifying the sparse structure in the regression weight $\vbeta^*$.
In \cite{Belloni-Chernozhukov11}, the authors studied the Lasso penalized quantile regression problem and proposed a consistent estimator at a near-oracle rate. 
The authors of \cite{wang2013l1} studied the estimator for the $\ell_1$ penalized LAD regression problem and showed that their estimator could achieve near-oracle risk performance.
Later, in \cite{wu2009variable,fan2014adaptive,fan2014strong}, the authors further considered using adaptive weighted Lasso or folded concave penalty to improve the estimation performance of the sparse LAD regression problem.
Estimating $\vbeta^*$ for a sparse LAD regression typically requires more effort than solving a sparse penalized OLS problem due to the non-smoothness property of the median loss.
The existing works proposed methods of interior-point algorithm \cite{koenker2005frisch}, greedy coordinate descent algorithm \cite{wu2008coordinate,peng2015iterative}, subgradient descent algorithm \cite{wang2017distributed}, majorization-minimization algorithm \cite{hunter2000quantile} or ADMM algorithm \cite{Gu-Fan-Kong18}, etc,  to solve the sparse LAD regression problem. 
However, these methods either lack the theoretical convergence guarantee or suffer a slow convergence rate.
Furthermore, none of these works could provide the guarantee of preserving data privacy.
In this work, we aim to study the efficient algorithm for sparse LAD regression problems and also to establish the theoretical guarantee of estimation consistency, algorithm convergence, and data privacy.

\subsection{Differential Private Regression}

Differential privacy (DP) \cite{dwork2014algorithmic} is a canonical metric for data privacy analysis. 
Under the DP framework, the privacy is measured by how significant the distribution of query outcome changes when we only replace one single sensitive data in the dataset.
Mathematically, the definition of DP can be stated as following:
\begin{defn}[$(\epsilon, \delta)$-Differential Privacy]\label{Def: dp}
Given any two adjacent data sets $\D,\D' \in \D^n$ differing by a single element, a randomized mechanism $\Mcc: \D^n \rightarrow \Rb^d$ provides $(\epsilon, \delta)$-differential privacy if it
produces response in any measurable set $A$ with probability 
\begin{align}
    \mathbb{P}[\Mcc(\D) \in A] \leq e^{\epsilon}\cdot P[\Mcc(\D') \in A] + \delta.
\end{align}
\end{defn}
Typically, the literature achieves the $(\epsilon, \delta)$-differential privacy by perturbing the query mechanism with a certain type of random noise, such as the Gaussian or Laplacian noise.
The scale of injected noise is jointly determined by the given privacy budget $\epsilon$ the $\ell_2$ sensitivity of the query function:
\begin{defn}[$\ell_2$-Sensitivity]\label{Def: l2 sensitivity}
The $\ell_2$-Sensitivity is defined as the maximum change in the $\ell_2$-Norm of the function value $f(\cdot)$ on two adjacent datasets $\D$ and $\D^\prime$: 
\begin{align}
   \Delta_2(f) = \max_{\D,\D^\prime} \|f(\D)-f(\D^\prime)\|_2, 
\end{align}
where $\D$ and $\D^\prime$ are two adjacent datasets that differ by only one element.
\end{defn}
So far, following the core definition of DP, various privacy concepts and perturbation methods have been proposed, such as local DP \cite{duchi2018minimax}, f-DP \cite{zheng2021federated}, output/objective perturbation \cite{chaudhuri2011differentially}.
Thanks to these data privacy techniques, privacy-preserving machine learning has achieved enormous success in a large number of scientific and engineering areas. 

As we study the sparse linear regression model, thus, in following, we mainly review the recent developments of privacy-preserving sparse regression methods and put our work into a comprehensive comparison.
The authors of \cite{cai2021cost} studied the trade-off between statistical accuracy and privacy for sparse linear regression in both low- and high-dimensional cases.
They showed that based on $\ell_2$ norm measurement, the optimal private minimax risk is lower bounded by $\Omega(\sqrt{{s\log p}/{N}} + {s\log p}/{N \epsilon})$ for the high-dimensional case and $\Omega(\sqrt{p/N}+ {p}/{N\epsilon})$ for the low-dimensional case.
In \cite{wang2019differentially}, the authors studied the differential private sparse linear model from the perspective of algorithm design. 
To achieve the desired $(\epsilon, \delta)$-DP, they proposed the DP-IGHT algorithm leveraging recently developed $\ell_0$ hard-thresholding and gradient perturbation.
They showed that their algorithm enjoys a linear convergence rate and attains the lower bound as in \cite{cai2021cost}.
\cite{wang2019sparse} studied the sparse linear model under local differential privacy.
They first showed that the polynomial dependency on $p$ in the lower bound of estimation error can not be avoidable if we need to protect the whole data privacy, and provided a private algorithm named LDP-IHT for low-dimensional regression that achieves near-optimal upper bound as $O({p\log p \log n}/{N\epsilon^2})$. 
Then they established the results that when only responses' privacy needs to be protected, their LDP-IHT algorithm could achieve a better error upper bound as $O(\sqrt{s\log p} /\sqrt{N}\epsilon)$.
However, these works focused on the conventional OLS loss and did not consider the model robustness. 
\cite{wang2022differentially} studied differentially private $l_1$-norm linear regression without sparse constraint on the regression coefficient. Furthermore, they consider a truncated loss, which introduces an additional approximation error of $O(\sqrt{p/N})$, which is not negligible when $p$ is at the same order of $N$.
\cite{hu2022high} extended to enable sparsity recovery by using a truncated square loss.
In our work, we study the robust sparse linear regression problem.
Instead of an OLS loss, we consider the median loss for linear model fitting. To our best knowledge, differential private sparse LAD regression is still an open problem and there are few theoretical analyses on this problem.
Table \ref{tab: compare_literature} summarizes the related works. The last column ``Oracle rate" indicates whether the work proved that as the privacy budget $\epsilon$ goes to infinity, the statistical accuracy of the model weights would achieve the oracle rate $\sqrt{s \log p/N}$.
{\setlength\tabcolsep{0pt}
\begin{table}[!ht]
\small
\caption{Comparison of different ($\epsilon$, $\delta$)-DP algorithms for linear regression.}
\centering
\begin{threeparttable}
\begin{tabular*}{\linewidth}{@{\extracolsep{\fill}} l|cccccccccc}
\hline
\hline
\multirow{2}{*}{Method} & \multirow{2}{*}{Privacy-accuracy trade-off} & Sparsity & Algorithm not & \multirow{2}{*}{Robustness} & Non-smooth & Oracle \\
 &  & recovery & depending on $s$ & & loss & rate \\
\hline
\cite{cai2021cost} & $O\{s\log p \log N[\log(1/\delta)\log N]^{1/2}/N \epsilon\}^*$ & \CheckmarkBold & \XSolidBrush & \XSolidBrush & \XSolidBrush & \CheckmarkBold \\
\cite{wang2019differentially} & $O\{s[\log(1/\delta)\log p]^{1/2}/N \epsilon\}^*$ & \CheckmarkBold & \XSolidBrush & \XSolidBrush  & \XSolidBrush & \CheckmarkBold \\
\cite{wang2022differentially} & $O\{[s^2 \log N \log(1/\delta)/(N \epsilon)]^{1/2}\}^{'}$ & \XSolidBrush & -- & \CheckmarkBold  & \XSolidBrush & \XSolidBrush \\
\cite{hu2022high} & $O\{s^2 \log N \log p ^2 \log(1/\delta)/(N \epsilon)\}^{'}$ & \CheckmarkBold & \XSolidBrush & \CheckmarkBold  & \XSolidBrush & \XSolidBrush\\
Ours & $O\{[p\log(1/\delta)\log(N\epsilon)]^{1/2}/N \epsilon\}^*$ & \CheckmarkBold & \CheckmarkBold & \CheckmarkBold & \CheckmarkBold & \CheckmarkBold \\
  \hline
  \hline
\end{tabular*}
\begin{tablenotes}
    \footnotesize
\item[] $^*$ Trade-off measured on weights. $^{'}$ Trade-off measured on loss.
\end{tablenotes}
\end{threeparttable}
\label{tab: compare_literature}
\end{table}}

\section{Problem and Algorithm} \label{sec:algorithm}

\subsection{Problem Formulation}

The motivation of the paper is to develop a fast and privacy-preserving algorithm for the sparse and robust regression problem. More specifically, we consider the following classical linear model: 
\[\y = \X \vbeta^*+\e,\]
where $\y = (y_1, y_2, \cdots, y_N)^\top$ is the vector of responses, $\X = (\x_1,\x_2,\cdots,\x_N)^\top$ is the $N \times p$ feature matrix, $\x_i = (x_{1,i},\cdots,x_{p,i})^\top$ is a $p$-dimensional covariate vector, $\vbeta^* = (\beta_1^*,\cdots,\beta_p^*)^\top$ is the true sparse regression weight, and $\e = (e_1, e_2, \cdots, e_N)^\top$
is the vector of random errors. We define $S=\{1 \leq i \leq p: \beta_{i}^{*} \neq 0\}$ as the support of $\vbeta^*$ and $s = \abs{S}$ as the sparsity level.
Assume that the random error has a density function $f$ and is independent of the covariate vector.  
It is common to assume that there exists a positive constant $c_{\vbeta}$ such that for the true regression weight, $\|\vbeta^*\|_2 \leq c_{\vbeta}$ such as in \cite{cai2021cost}.
Moreover, we assume that the covariate $\x$ has bounded $\ell_2$ norm, which is common for sparse and robust regressions (\cite{wang2019differentially,cai2021cost}).

Note that we have quite a slack assumption for the random error, which allows heavy-tailed random errors, i.e., abnormal responses such that the variance could be infinite, such as Cauchy distribution. For such a heavy-tailed error, the least square based methods are not applicable since the finite second-moment assumption does not hold. Our error assumption is looser than the state-of-the-art privacy-preserving algorithms designed for high-dimensional linear regression such as \cite{wang2019differentially,cai2021cost,hu2022high}, where finite second or fourth moments of the error is required. 

To tackle the heavy-tailed noise problem, we consider the least absolute deviation regression, i.e., the regression weight $\bbeta^*$ could be obtained by minimizing the median loss,
\[
\arg\min_{\bbeta \in \R^p} \frac{1}{N} \|\y -\X^\top \vbeta\|_1.
\]
Furthermore, to pursue the sparsity structure of $\bbeta^*$, we adopt a regularization term on the weights and obtain the sparse and robust estimated weights as 
\begin{align} \label{eq:medianloss_l1}
\widehat{\vbeta} = \arg\min_{\bbeta \in \R^p}  \frac{1}{N} \|\y -\X^\top \vbeta\|_1 + \lambda\|\vbeta\|_1,
\end{align}
where $|\cdot|_1$ is the $\ell_1$-regularization term and $\lambda>0$ is the regularization parameter. \cite{wang2019differentially,cai2021cost,hu2022high} utilized the $\ell_0$ hard-threshold to choose non-zero weights. However, in the simulation, we show that the $\ell_1$-regularization performs more stable than the $\ell_0$ hard-threshold method.
In this work, we aim to develop a fast and privacy-preserving algorithm to solve the sparse LAD problem $\vbeta$ in (\ref{eq:medianloss_l1}).

\subsection{Proposed Algorithm}

In this section, we propose our FRAPPE algorithm to find the sparse and robust estimation for $\vbeta^*$. Note that the sparse LAD problem has a non-smooth loss as well as a non-smooth penalty term. To our knowledge, the doubly non-smooth loss function is typically solved by subgradient descent algorithms \cite{wang2017distributed,wang2019distributed}. However, subgradient descent algorithms suffer from a slow convergence rate. To speed up the computation, we consider transforming the median loss into a square loss.

Our FRAPPE algorithm has a double-loop structure. In the outer loop, we construct pseudo responses such that the minimizer of the mean square error with the pseudo responses can be seen as an update in the Newton-Raphson iteration to solve the non-smooth problem (\ref{eq:medianloss_l1}). Thus, the $\ell_1$ penalized LAD loss is transformed as $\ell_1$ penalized square loss. In the inner loop, we solve the transformed $\ell_1$ penalized square loss privately.

Given a data-driven initial regression weight, which will be introduced later, all the parameters are updated iteratively. Without loss of generality, we introduce the proposed updating steps in the $v$th loop.
At the $v$th loop, given the current estimation $\hvbeta_v$, we first estimate the density of the error $\{e_i\}_{i=1}^N$.
The density estimation can be easily achieved by kernel estimation 
\begin{align} \label{eq: kernel_estimation}
\widehat{f}_v(0)=\frac{1}{Nh_{v}}\sum_{i=1}^N K\Big(\frac{y_i-\x_i^\top\widehat{\vbeta}_{v}}{h_{v}}\Big),
\end{align}
where $\widehat{f}_v(0)$ denotes the estimated density evaluated at point zero, $K(\cdot)$ is a kernel function and $h_v$ is the bandwidth at $v$-th iteration. 
To guarantee differential privacy, we add Gaussian noise to $\widehat{f}_v(0)$:
\begin{align}\label{eq. kernel noise}
\widehat{f}^\prime_{v}(0) = \max\{\widehat{f}_{v}(0) + u_{\widehat{f}_{v}}, c_{f(0)}^{-1}\},
\end{align}
where $u_{\widehat{f}_{v}} \sim N(0,\sigma^2_{\widehat{f}_{v}})$ and $c_{f(0)}$ is some positive constant to ensure that $\widehat{f}^\prime_{v}(0)$ would be bounded away from zero.
Then, we construct the pseudo responses $\widetilde{\y}_{i,v}$ for the $v$th outer loop
\begin{align}\label{eq:pseudo_y}
\widetilde{y}_{i,v} = \x_i^\top \hvbeta_{v}-\big(\widehat{f}^\prime_{v}(0)\big)^{-1}(\ind{y_i \le \x_i^\top \hvbeta_{v}}-1/2).
\end{align}
With the pseudo responses $\widetilde{\y}_{i,v}$, we reformulate the estimation problem as an $\ell_1$ penalized least square estimation problem 
\[\tbeta_v=\arg\min_{\vbeta \in \Bc} H_v(\vbeta) + \lambda_v\|\vbeta\|_1,\]
where $\Bc = \{\vbeta \in \Rb^p : \|\vbeta\|_2 \leq c_{\vbeta}\}$ and $H_v(\vbeta) = \frac{1}{2N}\sum_{i=1}^N (\widetilde{y}_{i,v} -\x_i^\top\vbeta)^2$.
It has been shown that the solution $\tbeta_v$ can be seen as the $v$th update in the Newton-Raphson iteration to solve problem (\ref{eq:medianloss_l1}). We say that steps (\ref{eq: kernel_estimation}-\ref{eq:pseudo_y}) are in the outer loop to reformulate the median loss as a square loss.

In the inner loop iterations, we solve the reformulated penalized least square problem with privacy constraint. We adopt the method of differential private gradient descent with soft thresholding.
At $t$th inner loop, we first update the regression weight as  
\begin{align}\label{eq. inner_beta_soft}
\hvbeta_{v,t-\frac{1}{2}} 
 =  \text{soft}\big(\hvbeta_{v,t-1} -   \eta(\nabla H_v(\hvbeta_{v,t-1}) + \u_{\hvbeta_{v,t}},\lambda_v\eta\big),
 \end{align}
where $\u_{\hvbeta_{v,t}} \sim N(0,\sigma_{\hvbeta_{v,t}}^2 \bI_p))$ and the soft operator is defined as  $\text{soft}(g,\tau) := \text{sign}(g)(|g|-\tau)_+$ and $(a)_+ = a \cdot \ind{a>0}$.
Then we do the weight clipping with a positive constant $c_{\vbeta}$
\begin{align}\label{eq. weight_clip}
\hvbeta_{v,t} =  \frac{c_{\vbeta}}{\max\{c_{\vbeta},\|\hvbeta_{v,t-\frac{1}{2}}\|_2\}}  \cdot \hvbeta_{v,t-\frac{1}{2}}.
\end{align}
Note that, unlike most of the existing work with gradient clipping, we perform the clipping on model weight, which projects the weight into a compact set. According to our problem formulation, the true model weight is bounded by a constant $c_{\bbeta}$, i.e. $\|\bbeta\| \le c_{\bbeta}$, so the above clipping would not make estimation biased. 
We summarize our algorithm in Algorithm~\ref{Algorithm: GT-QRE}.

\begin{algorithm}[ht]
    \caption{Fast Robust And Privacy-Preserving Estimation (FRAPPE) algorithm for least absolute deviation regression
    }\label{Algorithm: GT-QRE}
\begin{algorithmic} [1]
\REQUIRE  Data $\{y_i,\x_i\}_{i=1}^N$, step size $\eta$, noise scales $\sigma^2_{\hvbeta_0}$, $\sigma^2_{\hfv}$, $\sigma^2_{\hvbeta_{v,t}}$.
\STATE Initialize regression weight $\widehat{\vbeta}_0$, e.g. \eqref{eq: initial}, 

and inject the noise $\widehat{\vbeta}_{1} = \widehat{\vbeta}_0 + \u_{\widehat{\vbeta}_0}$, where $\u_{\widehat{\vbeta}_0} \sim N(0,\sigma_{\widehat{\vbeta}_0}^2 \bI_p))$; 
\FOR{$v = 1,\cdots, V$}
	\STATE Estimate the kernel density \eqref{eq: kernel_estimation} and inject the noise \eqref{eq. kernel noise};
	\STATE  Construct the pseudo response \eqref{eq:pseudo_y} and set up the weight $\widehat{\vbeta}_{v,0} = \widehat{\vbeta}_{v}$;
	\FOR{$t = 1,\cdots, T$}
		\STATE Update the weight following~\eqref{eq. inner_beta_soft} and~\eqref{eq. weight_clip};
	\ENDFOR
	\STATE Set up the weight $\widehat{\vbeta}_{v+1} = \widehat{\vbeta}_{v,T}$;
\ENDFOR
\end{algorithmic}
\end{algorithm}

\begin{rem}
We would like to preserve data privacy in each step where the data are used, so it is necessary to add noise to all three steps. Although we add noise to all three steps, we reduce the total noise addition by performing a loss transformation. Note that, for the initialization, we only add noise once. For kernel estimation, we add noise to every outer loop, and the total number of outer loops is up to a constant, as given in Theorem 3. Thus, most noise is added in the inner loop gradient descent step. For the inner loop part, we transform the problem to a least square estimation problem, which can be solved at a linear speed. Compared to a subgradient-based algorithm, which has a sublinear convergence rate, our algorithm has fewer inner loop iterations. Therefore, our algorithm could achieve a better privacy estimation trade-off.
\end{rem}

\noindent{\textbf{Initialization:}}
In this part, we discuss how to find an initial weight $\widehat{\vbeta}_0$ for Algorithm \ref{Algorithm: GT-QRE}.
Instead of the full dataset $\{y_i,\x_i\}_{i=1}^{N}$, we suggest using a subsampled dataset $\Sc$ with size $n$.
As only the $n$ subsampled data are involved, it could speed up the computation for initialization and enhance data privacy with privacy amplification (\cite{balle2018privacy}).
We consider estimating $\widehat{\vbeta}_0$ by solving the elastic net penalized LAD regression problem 
\begin{align}
\label{eq: initial}
\widehat{\vbeta}_{0} = \arg\min \frac{1}{n} \sum_{(y_i,\x_i)\in \Sc} |y_i -\x_i^\top \vbeta| + \lambda_{01}\|\vbeta\|_1+\frac{\lambda_{02}}{2}\|\vbeta\|_2^2,
\end{align}
as there exist many algorithms and software packages for it \cite{yi2017semismooth,su2021elastic}.
Because we do not specify an algorithm for solving the elastic net penalized LAD regression problem, we use the output perturbation to guarantee DP.

\section{Main Theoretical Results} \label{sec:theory}
In this section, we will first show that our proposed FRAPPE method satisfies $(\epsilon,\delta)$-DP and then prove the statistical consistency of our proposed estimator.
The theoretical results also provide specific parameter selections for the proposed FRAPPE algorithm.
We start our analysis with the following assumptions.
\begin{assum}\label{assum:bounded_x}
There exists a positive constant $c_\x$ such that $\|\x\|_2 \leq c_\x$.
\end{assum}

\begin{assum}\label{assum:density_f0}
The error $\e$ follows the density function $f(\cdot)$, which is bounded and Lipschitz continuous (i.e.,$|f(x) - f(y)| \leq C_L|x-y|$ for any $x, y \in \Rb$ and some constant $C_L > 0$). Moreover, we assume that its derivative $f'(\cdot)$ is bounded and $f(0) > c_{f(0)} > 0$ for some constant $c_{f(0)}$.
\end{assum}

\begin{assum}\label{assum:kernel_bound}
Assume that the kernel function $K(\cdot)$ is integrable with $\int_{-\infty}^{\infty} K(u) \mathrm{d} u=1$. Moreover, assume that $K(\cdot)$ satisfies $K(u) = 0$ if $|u| > 1$ and $K(u) \leq B$ for a positive constant $B$. Further, assume $K(\cdot)$ is differentiable
and its derivative $K'(\cdot)$ is bounded.
\end{assum}

\begin{assum}\label{assum:dim}
The total sample size $N$ satisfies $N = \Theta(p^{a+1})$ for some $a > 0$. 
The subsample size $n$ satisfies $n = \Theta(p^{r})$ for some $0 < r < a+1/2$, and the sparsity level $s$ satisfies $s = O(p^b)$ for some $0 < b < a/3 + 1/6$.
\end{assum}

Assumption \ref{assum:bounded_x} is a regular algorithmic assumption. Assumption \ref{assum:density_f0} is a statistical assumption for the linear model, which assumes the smoothness of the density function $f(\cdot)$ of the potentially unbounded errors. 
Assumptions \ref{assum:kernel_bound}-\ref{assum:dim} are required for the statistical convergence.
The Assumption \ref{assum:kernel_bound} is a standard assumption for kernel estimation. 
Assumption \ref{assum:dim} states requirements for the order of dimension $p$, sample size $n$ to calculate the initial estimator and the sparsity level $s$. 
The conditions of subsample size $n$ and sparsity level $s$ are used to ensure that the proposed algorithm will achieve the near-oracle convergence rate after a finite number of outer loop iterations, as shown in equation \eqref{eq:iteration_V}.

\begin{rem}
We considered the sparse LAD problem with $p < N$, as in Assumption 4, the only requirement for dimension $p$ is that $N = \Theta(p^{a+1})$ for some $a > 0$. This requirement covers the scenario of $p \ll N$ as in the low-dimension regime. In our framework, we allow $p$ to be in the same order as $N$, as long as the transformed LSE problem is strongly convex. For example, in the context of image analysis, our framework requires that the input for the proposed method should have a pixel count lower than the total number of images, and the pixel count could be in the same order as the number of images $N$. For the case with $p > N$, we could add regularizers to make the problem strongly convex \cite{shalev2009stochastic,neel2021descent}. 
\end{rem}

\subsection{Privacy Guarantee}
In this part, we will show that our FRAPPE algorithm satisfies $(\epsilon,\delta)$-DP with the proposed three-stage noise injection in Theorem 1. 
The proof of Theorem 1 utilizes the $\ell_2$-sensitivity for each stage and a standard composition theorem of zero-
concentrated differential privacy (zCDP) \cite{bun2016concentrated}.
For the first stage, we achieve the privacy guarantee for the data-driven initial point $\hvbeta_0$ via the output perturbation, which does not depend on the initialization algorithm.
As we set up $\hvbeta_0$ with the elastic net penalized LAD loss, the output perturbation is differential private by leveraging the $\lambda_{02}$-strongly convexity of the loss \cite{chaudhuri2011differentially}.
Then for the second stage, the DP of kernel density estimation $\hf_0$ follows from the fact that the kernel function is bounded so that the $\ell_2$-sensitivity of the kernel estimation is at the order of $O(1/N)$. 
In the last stage, due to the bounded gradient $\nabla H_v(\hvbeta_{v,t})$, we adopt the gradient perturbation method to achieve differential privacy.
By properly selecting the injected noises, each stage achieves $(\epsilon/3,\delta/3)$-DP. Since we aim to protect data privacy over the three stages without loss of generality, we use the even split for simplicity. More experiments for the effect of split of privacy budget $\epsilon$ is given in the supplementary material.
Using the composition property of zCDP, we guarantee that our proposed algorithm is $(\epsilon,\delta)$-DP.

\begin{thm}\label{thm:dp_together}
Suppose Assumptions \ref{assum:bounded_x}-\ref{assum:kernel_bound} hold, then Algorithm \ref{Algorithm: GT-QRE} is $(\epsilon, \delta)$-DP with 
$\sigma^2_{\hvbeta_0} = \frac{24c_{\x}^2 \log(n/(N\delta))} {\epsilon^2 \lambda_{02}^2 N^2}$,
$\sigma^2_{\hfv} = \frac{24B^2\log(1/\delta)V}{\epsilon^2N^2h_{v}^2}$  
and $\sigma^2_{\hvbeta_{v,t}} = \frac{6G^2\log(1/\delta)TV}{\epsilon^2N^2}$,
where $G = 4c_{x}^2c_{\beta}+c_{f(0)}$.
\end{thm}

We can see that reducing the iteration numbers $V$ and $T$ can improve privacy. Thus, the fast convergence of the proposed FRAPPE algorithm improves not only computation efficiency but also data privacy.

\subsection{Statistical Accuracy}

In this section, we provide the statistical convergence rate for the proposed estimator $\hvbeta_{v+1}$.
First, we show the rate of $\hvbeta_{2}$ after one iteration of the outer loop. 

\begin{thm} \label{thm:accuracy_one_round}
Suppose that Assumptions \ref{assum:bounded_x}-\ref{assum:dim} hold. Let $a_{1,N}=\sigma_{\widehat{\vbeta}_0}\sqrt{p}\log N+\sqrt{s\log p/n}$ and take $\lambda_{1}=C_{0}\left(\sqrt{\frac{\log p}{N}}+a_{1,N}(\sigma_{\widehat{\vbeta}_0}\sqrt{p}\log N+\sqrt{s\log p/n})\right)$
with the constant $C_{0}$. For the estimator $\hvbeta_{2}$ obtained after $T$ iterations of perturbed gradient descent (\ref{eq. inner_beta_soft}), if the learning rate is $1/2L$, and the number of iterations is $T = O(\log_{(1-\mu/2L)}\big(\frac{pG^2\log(1/\delta)}{N^2\epsilon^2}\big))$, we have 
\begin{align} \label{eq:accuracyini}
\|\hvbeta_2 - \vbeta^*\|_2 = O_{\mathbb{P}} & \left( a_{1,N}\sqrt{s}\left(\sigma_{\widehat{\vbeta}_0}\sqrt{p}\log N+\sqrt{s\log p/n}\right) +b_N+\sqrt{\frac{s\log p}{N}}\right), 
\end{align} 
where $b_N = \sqrt{p\log(1/\delta)\log(N\epsilon)}G/(N \epsilon)$.
\end{thm}

In the right-hand side of (\ref{eq:accuracyini}), the first term is from the error of the pseudo response construction $\widetilde{y}_{i,1}$ and the kernel density estimation $\widehat{f}^\prime_{1}(0)$. The second term $b_N$ is introduced by the perturbed gradient descent method. The third term is the error bound of the sparse least square estimation. 

\begin{rem}
The error bound of the initial estimator is $\sqrt{s\log p/n}$. 
Note that $\sigma_{\widehat{\vbeta}_0} = O(1/N)$. By Assumption 4, we have $\sigma_{\widehat{\vbeta}_0}\sqrt{p}\log N = O(\log p/p^{a+1/2})$ and $\sqrt{s\log p/n} = \Theta(\sqrt{\log p/p^{(1-r)b}})$, which implies $\sigma_{\widehat{\vbeta}_0}\sqrt{p}\log N$ is smaller than $\sqrt{s\log p/n}$. Similarly, we have $a_{1,N}\sqrt{s} = \Theta(\log p/p^{a+1/2-rb/2} + \sqrt{\log p/p^{(1-2r)b}})$, which is smaller than one. Therefore, the initial error $\sqrt{s\log p/n}$ gets shrunk. The term $b_N = O(\sqrt{\log p / p^{2a+1}})$ and the last term $\sqrt{s\log p/N} = O(\sqrt{\log p/p^{a-rb+1}})$ are in an smaller order of $\sqrt{s\log p/n}$. 
\end{rem}

By recursively applying Theorem \ref{thm:accuracy_one_round}, we provide the convergence rate for $\bhbeta_{v+1}$. 
Let us define, for $1\leq v\leq V$,
\begin{align*}
a_{v,N}= & \sqrt{\frac{s \log p}{N}}+b_N +\sqrt{s}(\sigma_{\widehat{\vbeta}_0}\sqrt{p}\log N+\sqrt{s\log p/n})^{v}.
\end{align*}
Then, we have the rate for $\bhbeta_{v+1}$ shown in the following theorem.
\begin{thm} \label{thm:accuracyv}
Assume that the initial estimator $\bhbeta_{0}$ satisfies $\|\bhbeta_{0}-\bbeta^{*}\|_{2}=O_{\mathbb{P}}(\sqrt{s\log p/n})$. Let $h_{v}=\Theta(a_{v,N})$ for $1\le v\le V-1$, and take
	\begin{align*}
	\lambda_{v,N}=C_{0}\left(\sqrt{\frac{\log p}{N}}+a_{v,N}(\sigma_{\widehat{\vbeta}_0}\sqrt{p}\log N+\sqrt{s\log p/n})\right)
	\end{align*} 
with $C_{0}$ being a sufficiently large constant. Under Assumptions \ref{assum:bounded_x}-\ref{assum:dim}, with inner loop learning rate $1/2L$ and inner loop iteration $T = O(\log_{(1-\mu/2L)}\big(\frac{pG^2\log(1/\delta)}{N^2\epsilon^2}\big))$, we have
	\begin{align*}
	 \|\bhbeta_{v+1}-\bbeta^{*}\|_{2} = O_{\mathbb{P}} & \left( \sqrt{\frac{s \log p}{N}}+b_N+ \sqrt{s}(\sigma_{\widehat{\vbeta}_0}\sqrt{p}\log N+\sqrt{s\log p/n})^{v+1}\right),
	\end{align*}	
where $b_N = \sqrt{p\log(1/\delta)\log(N\epsilon)}G/(N \epsilon)$.
\end{thm}

\begin{rem}
We give a fast way to find an initial estimator that satisfies the assumption of initialization in Theorem 3. The error bound of initialization is $\sqrt{s \log p/n}$, where $n$ could be much smaller than $N$. Then, Theorem 3 shows that under Assumption \ref{assum:dim}, after a constant number of iterations
\begin{align}
\label{eq:iteration_V}
V\ge\frac{2\log(N/\log p)}{\log(n/s\log p)},
\end{align}
the estimation accuracy will be improved and eventually achieve a rate $\sqrt{s \log p/N}+b_N$, and $\sqrt{s \log p/N}$ matches the oracle convergence rate up to a logarithmic factor and the term $b_N$ is due to the privacy accuracy trade-off. Compared with the initialization, the error bound is almost improved by a rate of $\sqrt{n/N}$.
\end{rem}

\begin{rem}
The privacy cost of our proposed estimator is of the order of $O(\sqrt{p})$, which polynomially depends on the dimensionality $p$.
This order is intrinsically caused by the soft thresholding for the $\ell_1$ penalty.
Under the worse scenario, the soft thresholding operator would keep all the elements of $\hat{\bbeta}_{v, t-\frac{1}{2}}$ as non-zero, so that the injected $p$-dimensional noises would be accumulated during the estimation.
Some previous works achieved the privacy cost at the order of $O(s\log p)$ by using peeling algorithm \cite{cai2021cost} or hard thresholding operation \cite{wang2019differentially}, which only remain a fixed number of the largest elements in regression weight during optimization. 
However, these methods require prior knowledge of the sparsity level and extra convexity condition on the loss, which cannot easily be extended to our case. 
As it is a non-trivial task to improve the privacy cost, we would like to leave it as a future study.
\end{rem}

\section{Numerical Evaluation} \label{sec:simulation}

In this section, we empirically evaluate the statistical and computational efficiency of the proposed FRAPPE algorithm on both synthetic data and real datasets.

\subsection{Synthetic Data Experiments}

\noindent{\textbf{Data genaration:}}
We generate data from the linear model: $y_i = \x_i^\top \vbeta^* + e_i$, $i = 1,\cdots,N$, where $\x_i$ is a $p$-dimensional vector, $\vbeta^* = (\beta_1^*,\cdots,\beta_p^*)^\top$ is the true regression weight, and $e_i$ is the random noise. The vectors $\{\x_i\}$ are drawn i.i.d. from multivariate normal distribution with mean zeros and covariance matrix $\Sigma$, which is constructed as $\Sigma_{ij} = 0.1^{|i-j|}$, for $1 \leq i,j \leq p$. 
Let $s$ be the sparsity level, and then the true regression weight is set as $\vbeta^* = s^{-1}10 \cdot (1,2,\cdots,s,0,,\cdots,0)$.
The random noise $e_i$ is i.i.d. sample from some given distributions.

\noindent \textbf{Compared algorithms:}
We compare the proposed FRAPPE algorithm with five other state-of-art algorithms designed for private sparse linear regression:
\begin{itemize}
\item SgpLAD: using subgradient perturbation descent (SGP) algorithm to solve the $\ell_1$ penalized least absolute deviation loss (LAD). The subgradient descent method is obtained from \cite{wang2017distributed}. To achieve $(\epsilon,\delta)$-differential privacy, we add random Gaussian noise to the subgradient in each iteration of Algorithm 1 in \cite{wang2017distributed}.
\item GpLASSO: using gradient perturbation (GP) to minimize $\ell_1$ penalized square loss. The minimizer of $\ell_1$ penalized square loss is known as the Lasso estimator. To keep privacy, random Gaussian noise is added to the gradient in each iteration when solving the Lasso problem.
\item DPSLR: Differentially Private Sparse Linear Regression given in Algorithm 4.2 of \cite{cai2021cost}. This estimator is designed to minimize the square loss. To recover the sparsity of the regression weights, the ``peeling" algorithm, proposed by \cite{dwork2018differentially}, is used to select $s$ weights with the top-$s$ largest absolute values in a differentially private way.
\item HtPSLR: Heavy-tailed Private Sparse Linear Regression given in Algorithm 3 of \cite{hu2022high}. First, they truncate the original data at a given level to avoid heavy-tailed data. Then, gradient descent and "peeling" are performed to obtain a sparse regression weight.
\item DPIGHT: Differentially Private Iterative Gradient Hard Thresholding proposed by \cite{wang2019differentially} and the specific algorithm for the sparse linear regression problem is given in section 5.2 of \cite{wang2019differentially}. This estimator aims to minimize the square loss. In each iteration, the weights are updated by perturbed gradient and then  perform the hard thresholding operator.
\end{itemize}

\noindent \textbf{Simulation setup:}
The initial estimator is obtained by solving (\ref{eq: initial}) with a randomly selected sample with size $n=200$. 
We use a biweight kernel function 
\[
K(x)=\left\{\begin{array}{ll}
0, & x \leq-1, \\
-\frac{315}{64} x^6+\frac{735}{64} x^4-\frac{525}{64} x^2+\frac{105}{64}, & -1 \leq x \leq 1, \\
0, & x \geq 1
\end{array}\right.
\]
It is easy to verify that the biweight kernel function satisfies Assumption 3. Other common kernel functions, such as uniform, triweight, and Epanechnikov, also satisfy Assumption 3 and share similar experimental results, which are shown in the experiment for the effect of kernel in the supplement material.
For the choice of bandwidth, as shown in Theorem \ref{thm:accuracyv}, it should have the same order as $a_{v,N}$, and thus we set the bandwidth as $h_v =\sqrt{\frac{s \log N}{N}}+s^{-1 / 2}\left(0.9\right)^{(v+1) / 2}$.
For the $(\epsilon, \delta)$-differential privacy, we set $\delta=10^{-3}$.
Additionally, we set the inner loop iteration $T=50$ and the outer loop iteration $V=10$.
To give a fair comparison, we keep the same number of total iterations for other methods, i.e., the total number of iterations for other methods is 500. 

We specify 20 values for the regularization parameter $\lambda$ according to the input data for the soft-thresholding based methods FRAPPE, SgpLAD, and GpLASSO. Also, 20 values for the sparsity level $s$ are specified for the hard-thresholding based methods DPSLR, HtPSLR, and DPIGHT. The $\lambda$ and $s$ with the smallest Bayesian information criterion (BIC) value are chosen for each algorithm, respectively.


\noindent \textbf{Performance measurements:}
In the experiment with the synthetic dataset, we evaluate the performance of the six methods in terms of the following two measurements:
\begin{enumerate}
    \item Mean square error (MSE) of the estimated regression weights to examine the model accuracy.
    \item $F_1$-score to evaluate the sparsity recovery. The $F_1$-score is defined as 
    \[F_{1}=\left(\frac{\text {recall}^{-1}+\text {precision}^{-1}}{2}\right)^{\!-1},\]
    where recall is the fraction of correctly identified non-zero weights over true sparsity and precision is the fraction of correctly identified non-zero weights over total identified non-zero ones. Note that the $F_1$ score ranges between 0 and 1, and the closer to 1, the better the sparsity recovery.
\end{enumerate}

\noindent \textbf{Effect of Heavy-Tailed Noises:}

\begin{wrapfigure}{R}{0.40\textwidth}
\centering
\includegraphics[width=.40\textwidth]{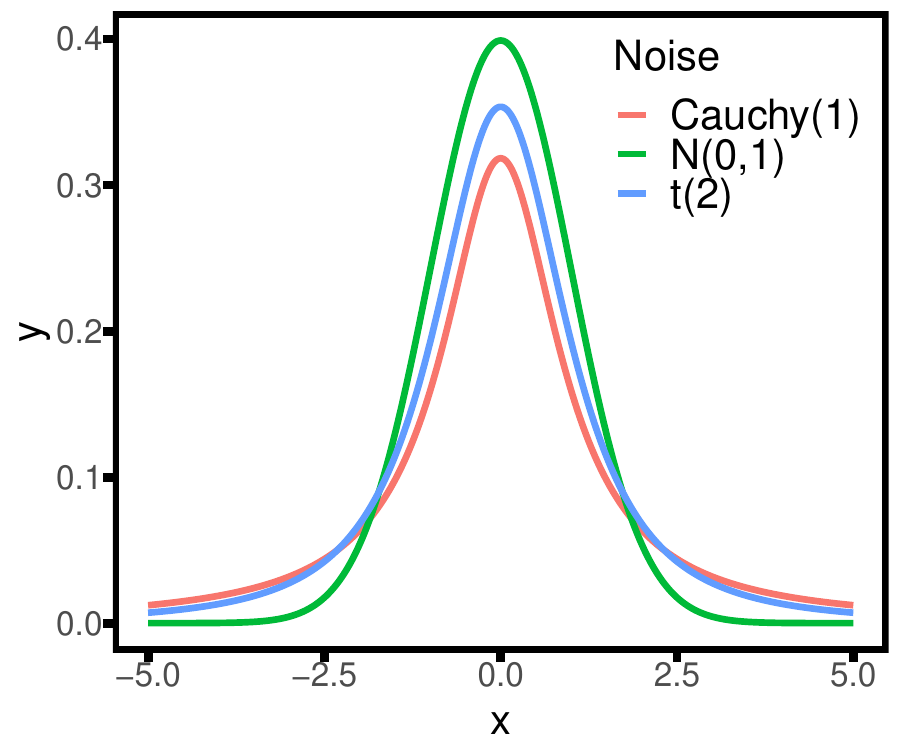}
\caption{The probability densities for the three noise distributions. $\mathrm{Cauchy}$ distribution has a heaviest tail and N(0,1) has the lightest tail.}
\label{fig:noise_density}
\vspace{-.3in}
\end{wrapfigure}

The proposed FRAPPE algorithm is designed for data contaminated with heavy-tailed noises and extreme values, which are pervasive in the real world. In this set of experiments, we aim to compare the practical performance of the proposed FRAPPE algorithm with the five other state-of-art differentially private sparse linear regression algorithms under different choices of noise distributions. We fix the dimension of $\x_i$ at $p=100$, sparsity level $s = 10$, and privacy budget $\epsilon = 0.5$. We consider the following three noise distributions:

\begin{itemize}
    \item Normal: the noise $e_i \sim N(0,1)$. Normal distribution has a light tail and finite moments.
    \item Student's t: the noise $e_i \sim t(2)$. Student's t distribution has a tail between Normal and $\mathrm{Cauchy}$ distributions, since $t(1)$ is equivalent to $\mathrm{Cauchy}$ distribution and $t(\infty)$ is Normal distribution.
    \item $\mathrm{Cauchy}$: the noise $e_i \sim \mathrm{Cauchy}(0,1)$. $\mathrm{Cauchy}$ distribution has a heavy tail and infinite variance.
\end{itemize}
Figure \ref{fig:noise_density} shows the probability densities for the three noise distributions. We can easily observe that $\mathrm{Cauchy}$ distribution has a heavier tail than t(2), and N(0,1) has the lightest tail behavior.

Table \ref{tab: noise} shows the MSE and $F_1$-score of all methods under three different noise distributions. Remember that FRAPPE and SgpLAD are based on LAD loss, and the other four methods are based on square loss. For Normal noise, the best methods are FRAPPE and DPIGHT, which indicates that under light-tailed noise, the LAD loss and square loss have similar performance. SgpLAD is worse than FRAPPE due to its slow convergence, especially when the data size decreases. DPIGHT outperforms GpLASSO, DPSLR, and HtPSLR because its iterative gradient hard thresholding procedure is more efficient than soft thresholding and ``peeling" procedure for sparse learning. 

For the Student's t(2) noise, FRAPPE, HtPSLR, and DPIGHT have similar performance. Because HtPSLR and DPIGHT employ data trimming to adopt abnormal observations and this method works for noise with finite second moment. GpLASSO and DPSLR perform worst as they are not designed for heavy-tailed noise. 

For heavy-tailed $\mathrm{Cauchy}$ noise, the proposed FRAPPE method has the best performance among all methods. The heavy-tailed noise deteriorates the estimation performance of the methods based on square loss. Because Heavy-tailed noise breaks the finite second-moment assumption for least square methods. HtPSLR shrunk the original data to avoid extreme values, which leads to a better performance than the no-shrinkage algorithm DPSLR, but HtPSLR still needs a finite second moment assumption that does not hold for $\mathrm{Cauchy}$ noise. More precisely, FRAPPE reduces at least about 63.6\% of MSE of the square loss based methods.

Note that as the total data size $N$ increases, the MSEs of all methods decrease, and as the dimension of the covariates increases, the MSEs of all methods increase, which matches the oracle convergence error $\sqrt{s \log p / N}$.  Moreover, when the data size $N$ gets larger or the dimension $p$ gets smaller, the MSE of SgpLAD increases much faster than that of FRAPPE. Because SgpLAD converges much slower than FRAPPE, and we will show that in the later section.

Thus, the consistent performance of the FRAPPE algorithm confirms that the proposed FRAPPE is robust to both light-tailed and heavy-tailed noises.

\begin{table*}[!ht]
\small
\caption{MSE and $F_1$ score for method comparison. We report results under three different noise distributions and sample sizes $N$. We fix privacy budget $\epsilon$ as 0.5, sparsity level $s$ as 10, and dimension $p$ as 100. The bold values under the same setting (row) are the two smallest MSEs.}
\label{tab: noise}
\centering
\begin{tabular}{l|c|rr|rr|rr|rr|rr|rr}
\hline
\hline
\multirow{2}{*}{Noise} & \multirow{2}{*}{$N$} & \multicolumn{2}{c|}{FRAPPE} & \multicolumn{2}{c|}{SgpLAD} & \multicolumn{2}{c|}{GpLASSO} & \multicolumn{2}{c|}{DPSLR} & \multicolumn{2}{c|}{HtPSLR} & \multicolumn{2}{c}{DPIGHT} \\
 & & MSE & $F_1$ & MSE & $F_1$ & MSE & $F_1$ & MSE & $F_1$ & MSE & $F_1$ & MSE & $F_1$ \\
 \hline 
\multirow{3}{*}{Normal}
    &  2000 & \textbf{0.05} & 0.91 & 6.15 & 0.64 & 1.76 & 0.20 & 2.25 & 0.75 & 0.10 & 0.93 & \textbf{0.05} & 0.98 \\ 
    &  5000 & \textbf{0.01} & 0.92 & 0.08 & 0.91 & 0.32 & 0.21 & 0.87 & 0.85 & 0.03 & 0.96 & \textbf{0.01} & 0.99 \\ 
    & 10000 & \textbf{0.01} & 0.95 & \textbf{0.01} & 0.96 & 0.09 & 0.21 & 0.90 & 0.83 & 0.02 & 0.96 & \textbf{0.01} & 1  \\
 \hline 
\multirow{3}{*}{Student's t}
    &  2000 & \textbf{0.31} & 0.96 & 5.11 & 0.63 & 1.88 & 0.22 & 2.31 & 0.75 & 0.37 & 1.00 & \textbf{0.22} & 0.99 \\ 
    &  5000 & \textbf{0.18} & 0.96 & 0.24 & 0.96 & 0.99 & 0.63 & 0.88 & 0.92 & 0.22 & 1.00 & \textbf{0.13} & 1.00 \\ 
    & 10000 & \textbf{0.12} & 0.96 & \textbf{0.12} & 0.97 & 0.44 & 0.91 & 0.93 & 0.96 & 0.17 & 1.00 & \textbf{0.10} & 1.00 \\ 
 \hline 
\multirow{3}{*}{$\mathrm{Cauchy}$}
    &  2000 & \textbf{0.44} & 0.99 & 5.91 & 0.66 & 5.50 & 0.55 & 2.83 & 0.82 & 1.45 & 0.96 & \textbf{1.19} & 0.91 \\ 
    &  5000 & \textbf{0.23} & 0.98 & \textbf{0.32} & 0.99 & 2.39 & 0.96 & 1.08 & 0.99 & 0.95 & 0.99 & 0.85 & 0.92 \\ 
    & 10000 & \textbf{0.15} & 0.98 & \textbf{0.14} & 0.98 & 1.08 & 0.98 & 0.96 & 1.00 & 0.62 & 1.00 & 0.51 & 0.93 \\ 
  \hline
  \hline
\end{tabular}
\end{table*}

\begin{table*}[!ht]
\small
\caption{MSE and $F_1$ score for method comparison. We report results under three different noise distributions and dimensions $p$. We fix privacy budget $\epsilon$ as 0.5, sparsity level $s$ as 10, and data size $N$ as 5000. The bold values under the same setting (row) are the two smallest MSEs.}
\label{tab: noise_p}
\centering
\begin{tabular}{l|c|rr|rr|rr|rr|rr|rr}
\hline
\hline
\multirow{2}{*}{Noise} & \multirow{2}{*}{$p$} & \multicolumn{2}{c|}{FRAPPE} & \multicolumn{2}{c|}{SgpLAD} & \multicolumn{2}{c|}{GpLASSO} & \multicolumn{2}{c|}{DPSLR} & \multicolumn{2}{c|}{HtPSLR} & \multicolumn{2}{c}{DPIGHT} \\
 & & MSE & $F_1$ & MSE & $F_1$ & MSE & $F_1$ & MSE & $F_1$ & MSE & $F_1$ & MSE & $F_1$ \\
  \hline 
\multirow{3}{*}{Normal}
    &  50 & \textbf{0.01} & 0.91 & \textbf{0.01} & 0.96 & 0.16 & 0.34 & 0.85 & 0.83 & 0.03 & 0.93 & \textbf{0.01} & 0.99 \\ 
    &  100 & \textbf{0.01} & 0.91 & 0.09 & 0.92 & 0.32 & 0.20 & 0.83 & 0.82 & 0.02 & 0.95 & \textbf{0.01} & 1.00 \\ 
    & 200 & \textbf{0.02} & 0.90 & 0.25 & 0.96 & 0.65 & 0.12 & 0.85 & 0.81 & 0.03 & 0.94 & \textbf{0.01} & 0.99 \\ 
  \hline 
\multirow{3}{*}{Student's t}
    &  50 & \textbf{0.16} & 0.97 & \textbf{0.16} & 0.97 & 0.46 & 0.61 & 0.91 & 0.91 & 0.24 & 1.00 & \textbf{0.13} & 1.00 \\ 
    &  100 & \textbf{0.18} & 0.96 & 0.24 & 0.96 & 0.99 & 0.63 & 0.88 & 0.92 & 0.22 & 1.00 & \textbf{0.13} & 1.00 \\ 
    & 200 & \textbf{0.21} & 0.97 & 0.69 & 0.98 & 1.26 & 0.30 & 0.89 & 0.88 & 0.24 & 1.00 & \textbf{0.14} & 0.99 \\ 
 \hline 
\multirow{3}{*}{$\mathrm{Cauchy}$}
    &  50 & \textbf{0.19} & 0.98 & \textbf{0.21} & 0.98 & 1.19 & 0.81 & 1.07 & 0.99 & 0.98 & 0.99 & 0.71 & 0.95 \\ 
    &  100 & \textbf{0.22} & 0.99 & \textbf{0.27} & 0.98 & 2.32 & 0.96 & 1.06 & 0.99 & 0.95 & 0.99 & 0.81 & 0.92 \\ 
    & 200 & \textbf{0.25} & 0.99 & 1.05 & 0.95 & 2.49 & 0.69 & 1.24 & 0.96 & 1.04 & 0.98 & \textbf{0.88} & 0.91 \\ 
  \hline
  \hline
\end{tabular}
\end{table*}

\noindent \textbf{Effect of the Sparsity Level:}
In this set of experiments, we evaluate the performance of all algorithms under different levels of sparsity. The oracle convergence error is $\sqrt{s \log p / N}$, and thus the MSE should increase with the sparsity level. 

\begin{figure*}
\centering
\begin{tabular}{@{}ccc@{}}
\includegraphics[width=0.3\linewidth]{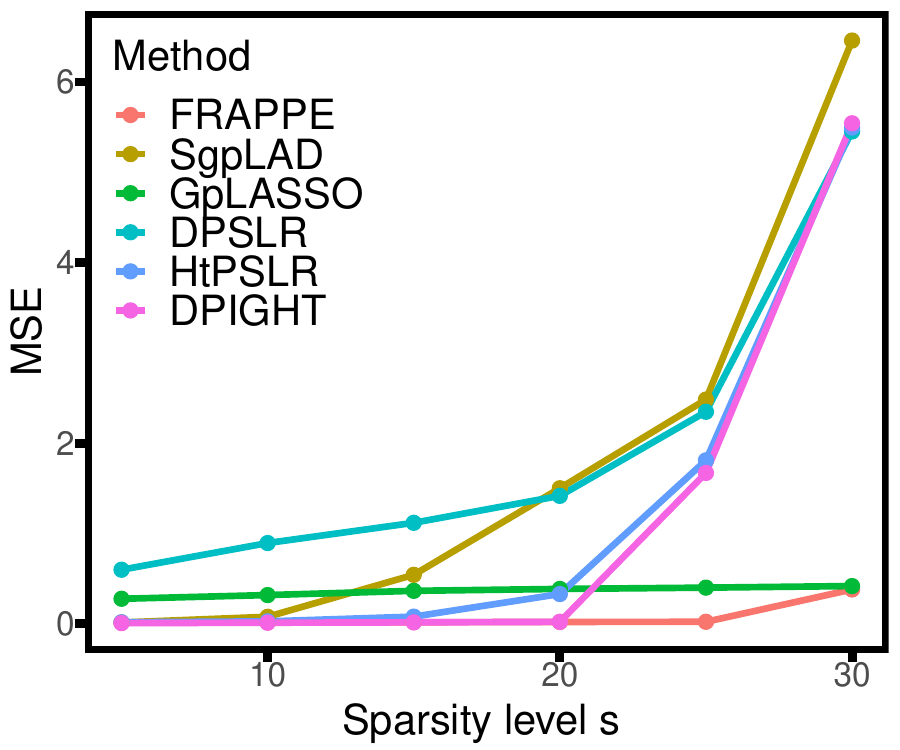} &
\includegraphics[width=0.3\linewidth]{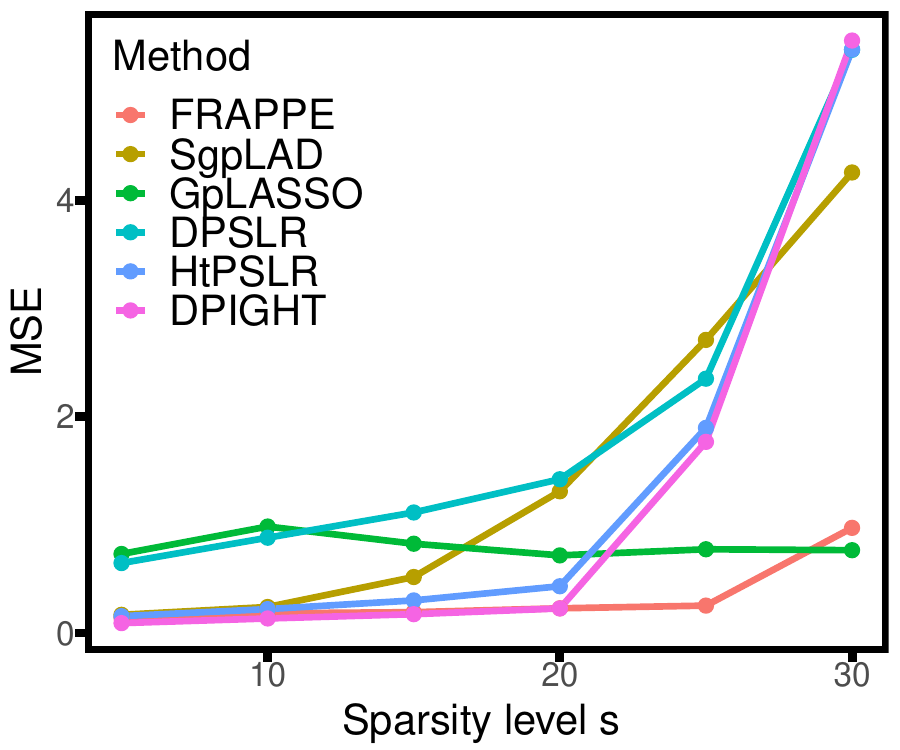} &
\includegraphics[width=0.3\linewidth]{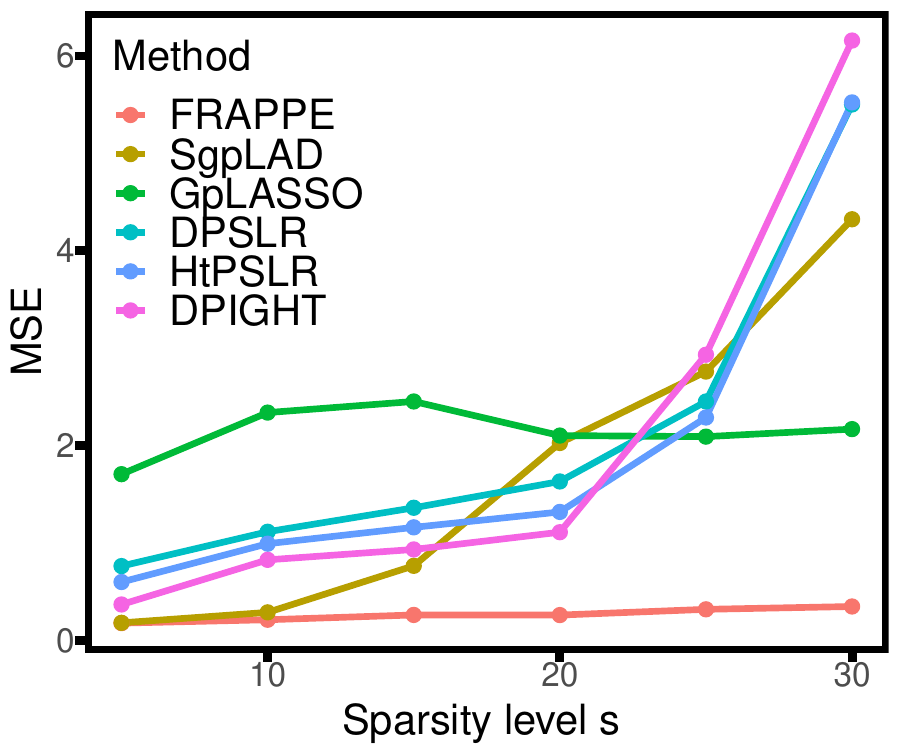} \\
{\small{(a) Normal noise}} & {\small{(b) Student's t noise}} & {\small{(c) $\mathrm{Cauchy}$} noise}
\end{tabular}
\caption{The MSE vs the sparsity level $s$ ranging from 1 to 30. The three figures are corresponding to different noise distributions under sample size $N=5000$, dimension $p=100$, and privacy budget $\epsilon=0.5$. Compared with the existing algorithms, our method achieved the best performance.}
\label{fig:sparse_mse}
\end{figure*}

We fix sample size $N=5000$, dimension $p=100$, and privacy budget $\epsilon = 0.5$. The results of MSE versus sparsity level $s$ under different noise distributions are reported in Figure~\ref{fig:sparse_mse}. We observe that the MSE increases as the sparsity level $s$ increases for all methods except for GpLASSO under heavy-tailed loss. Because the heavy-tailed data destroy the statistical convergence theory for GpLASSO, the GpLASSO does not follow the oracle property anymore. The MSE of SGPLAD has a faster increase rate than the oracle rate due to its slow convergence when $s$ gets larger. We can see that when the sparsity level gets larger than 20, the MSE of DPSLR, HtPSLR and DPIGHT have a faster increase rate than the oracle. This finding indicates the unstable performance of hard thresholding when the number of non-zero regression weights is large.  

Note that the proposed FRAPPE has the best performance among all the methods under all noise types and sparsity levels. Also note that the outer loop iteration numbers should increase with the sparsity $s$ to achieve the same convergence rate as shown in Theorem \ref{thm:accuracyv}. The results are based on 10 outer loop iterations, and hence, FRAPPE only requires a small number of outer loop refinements to achieve its theoretical statistical error. Thus, overall, FRAPPE performs consistently well in spite of light-tailed or heavy-tailed noise and different sparsity level.

\begin{figure*}
\centering
\begin{tabular}{@{}ccc@{}}
\includegraphics[width=0.3\linewidth]{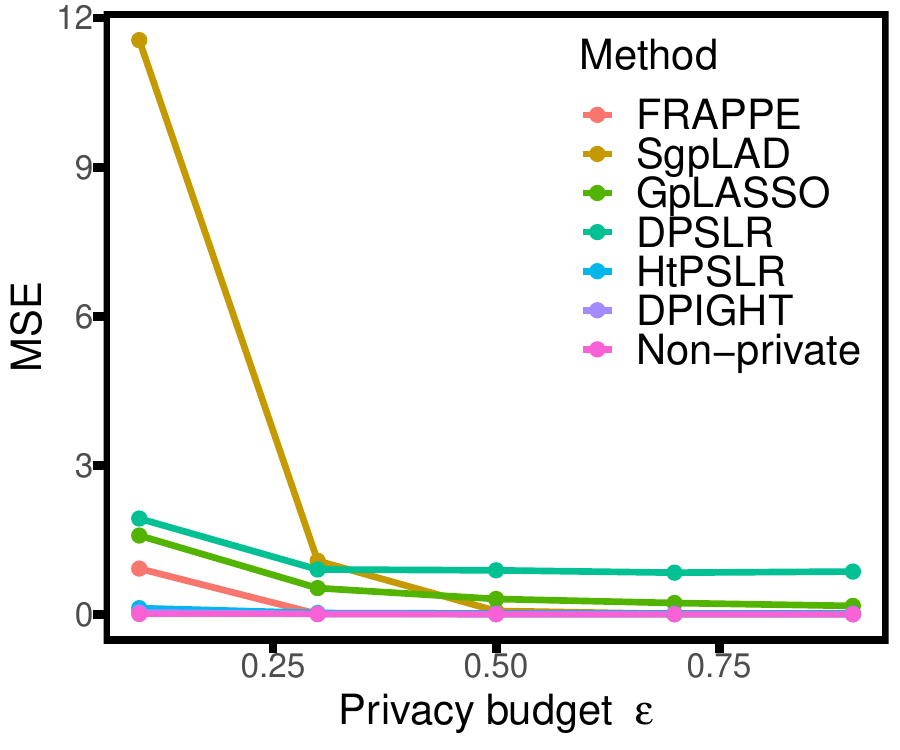} &
\includegraphics[width=0.3\linewidth]{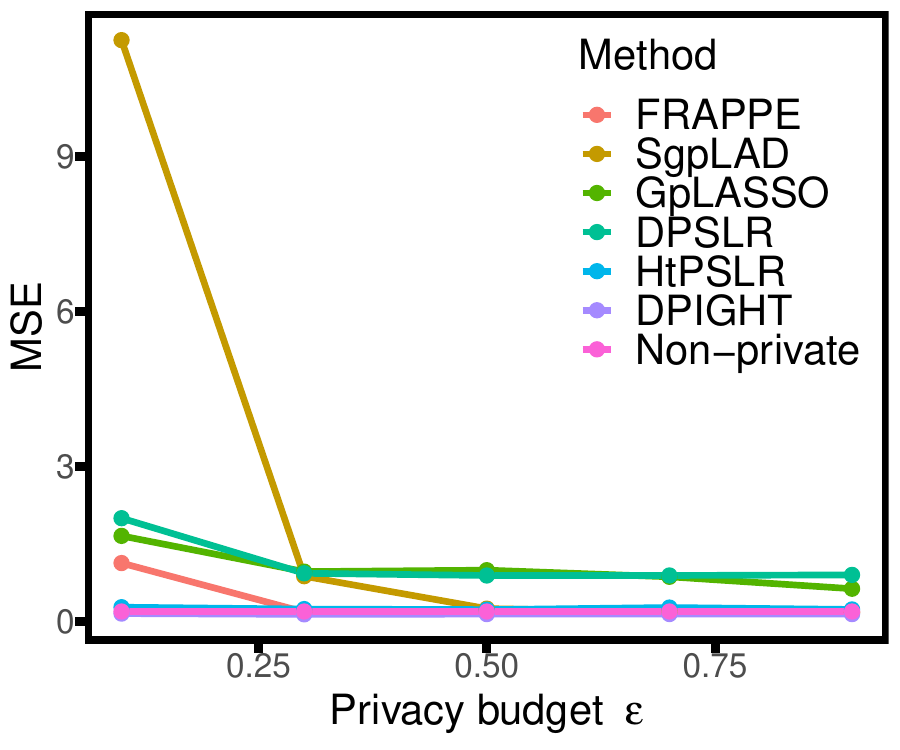} &
\includegraphics[width=0.3\linewidth]{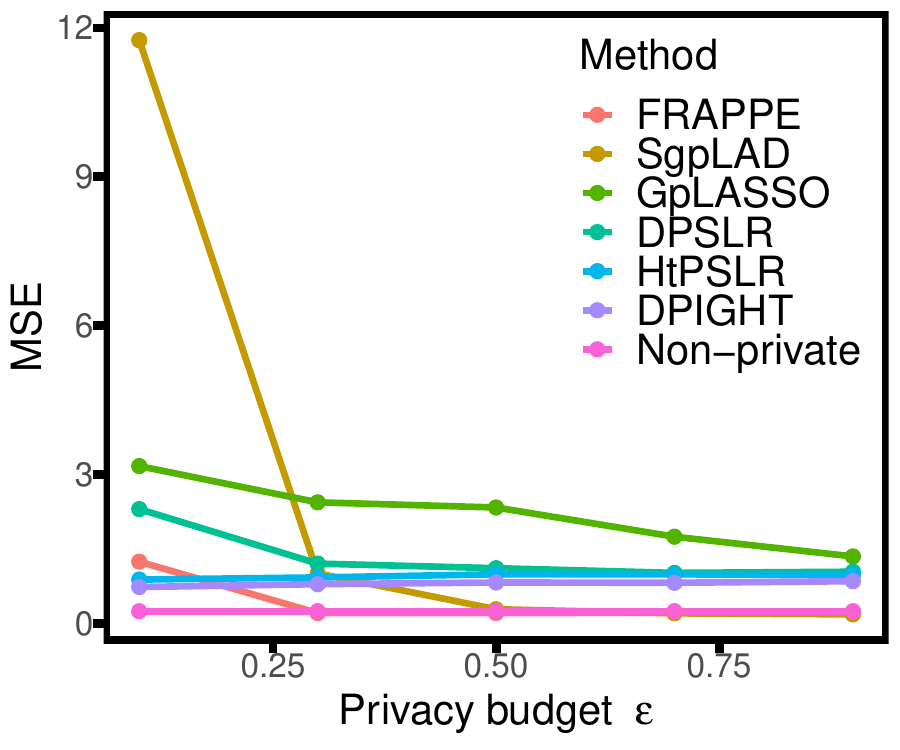} \\
{\small{(a) Normal noise}} & {\small{(b) Student's t noise}} & {\small{(c) $\mathrm{Cauchy}$} noise}
\end{tabular}
\caption{The MSE vs the privacy budget $\epsilon$ ranging from 0 to 1. The three figures are corresponding to different noise distributions under sample size $N=5000$, dimension $p=100$, and sparsity level $s=10$. Compared with the existing algorithms, our method achieved the best performance.}
\label{fig: epsilon}
\end{figure*}

\noindent \textbf{Effect of the Privacy Budget:}
In this set of experiments, we investigate the performance of all algorithms under different privacy budgets $\epsilon$. The differential private budget $\epsilon$ controls the privacy level, and the smaller value indicates a more stringent privacy constraint. We also compare the results obtained from the non-private version of the proposed FRAPPE algorithm (Non-private).

We fix sample size $N=5000$, dimension $p=100$, and sparsity level $s = 10$. The results of MSE versus privacy budget $\epsilon$ under different noise distributions are reported in Figure~\ref{fig: epsilon}. We can observe the privacy and statistical accuracy trade-off for all methods. For FRAPPE, the decrease speed of MSE becomes slower as increasing $\epsilon$, which is consistent with our theoretical result. As shown in Theorem \ref{thm:accuracyv}, for FRAPPE, the order of MSE is $O(\sqrt{s \log p /N}+\sqrt{p \log(N \epsilon)}/(N \epsilon))$. Note that the MSE of HtPSLR and DPIGHT changes slightly as the privacy budget $\epsilon$ changes. There is no statistical accuracy of the regression weights for HtPSLR. For DPIGHT, \cite{wang2019differentially} showed that the order of MSE is $O(\sqrt{s \log p /N}+s\sqrt{\log p}/(N\epsilon))$. The second term $s\sqrt{\log p}/(N\epsilon)$ shows the privacy accuracy trade-off and is smaller than FRAPPE one at the given setting in the experiment. However, except for the smallest $\epsilon$, FRAPPE performs the best among all privacy budget $\epsilon$ and different noise types.

\begin{figure*}
\centering
\begin{tabular}{@{}cc@{}}
\includegraphics[width=0.3\linewidth]{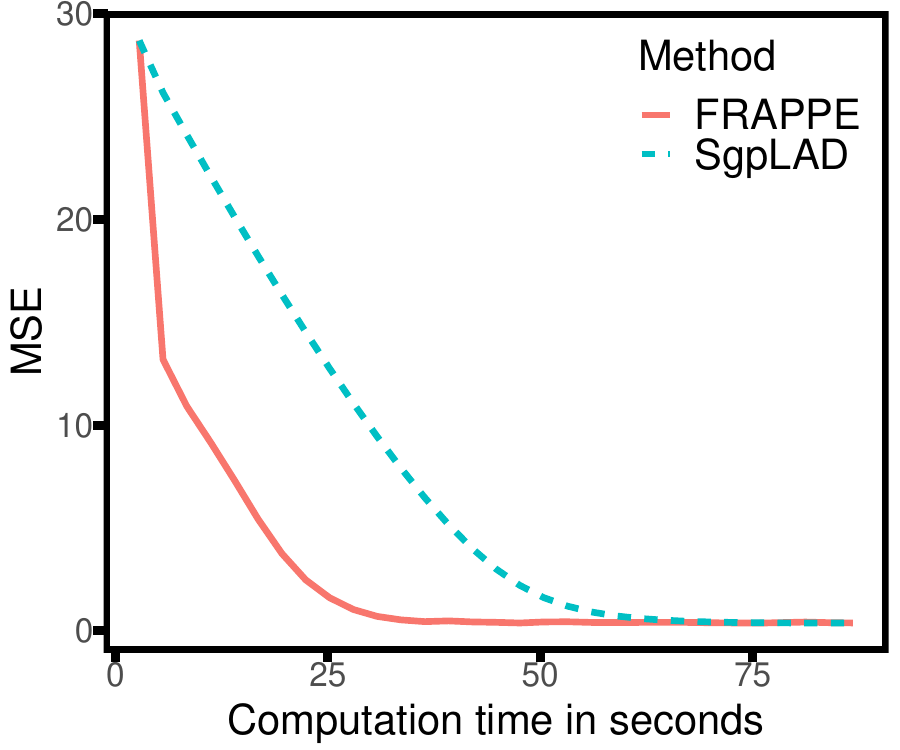} &
\includegraphics[width=0.3\linewidth]{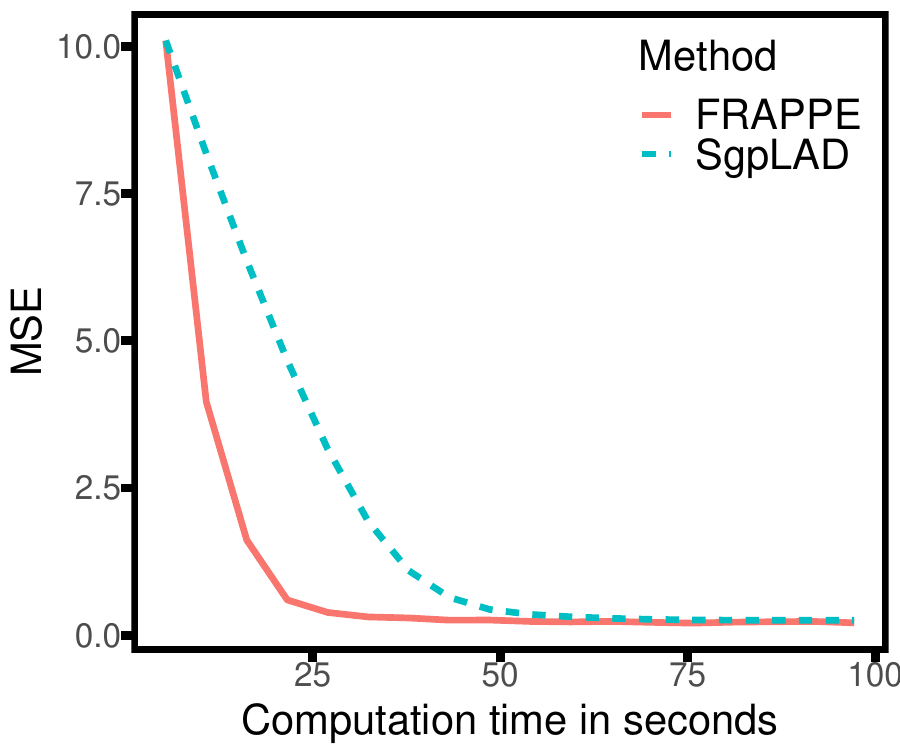} \\
{\small{(a) $N=2000$}} & {\small{(b) $N=5000$}}
\end{tabular}
\caption{The MSE vs computation time. We report the estimation MSE as computation time increases. The two figures correspond to different sample sizes $N=2000$ and $N=5000$ under $\mathrm{Cauchy}$ noise, sparsity level $s=10$, dimension $p=100$, and privacy budget $\epsilon=0.5$. Compared with the SgpLAD algorithm, our method performed better.}
\label{fig:time_mse}
\end{figure*}

\noindent \textbf{Computation Time Comparison:}
In this section, we further study the computation efficiency of our proposed estimator. The motivation of the proposed method is to shorten the computation time for the LAD problem. Hence, we compare the proposed FRAPPE algorithm with SgpLAD algorithm. We fix the privacy budget $\epsilon = 0.5$, sparsity level $s=10$, data size $N=5000$, and dimension $p=100$. In Figure \ref{fig:time_mse}, we report the MSE versus computation time in seconds of FRAPPE and SgpLAD with $\mathrm{Cauchy}$ noise. For the Normal and Student's t noise, the results are similar. We can observe that to achieve the same MSE, the computation time for FRAPPE is only about half of that for SgpLAD. 

Some more experiments evaluated the effect of inner loop and outer loop iterations, the effect of different splits of the privacy budget, the effect of the initial sample size, and the effect of different choices of the kernel functions for density estimation are shown in the supplementary material.

\subsection{Real Data Analysis}

In this section, we study the performance of the FRAPPE algorithm on two real datasets: the Communities and Crime dataset from the UCI Machine Learning Repository \cite{redmond2009crime} and the Ames Housing dataset, which is publicly available at Kaggle.

\begin{wrapfigure}{R}{0.40\textwidth}
\centering
\includegraphics[width=.40\textwidth]{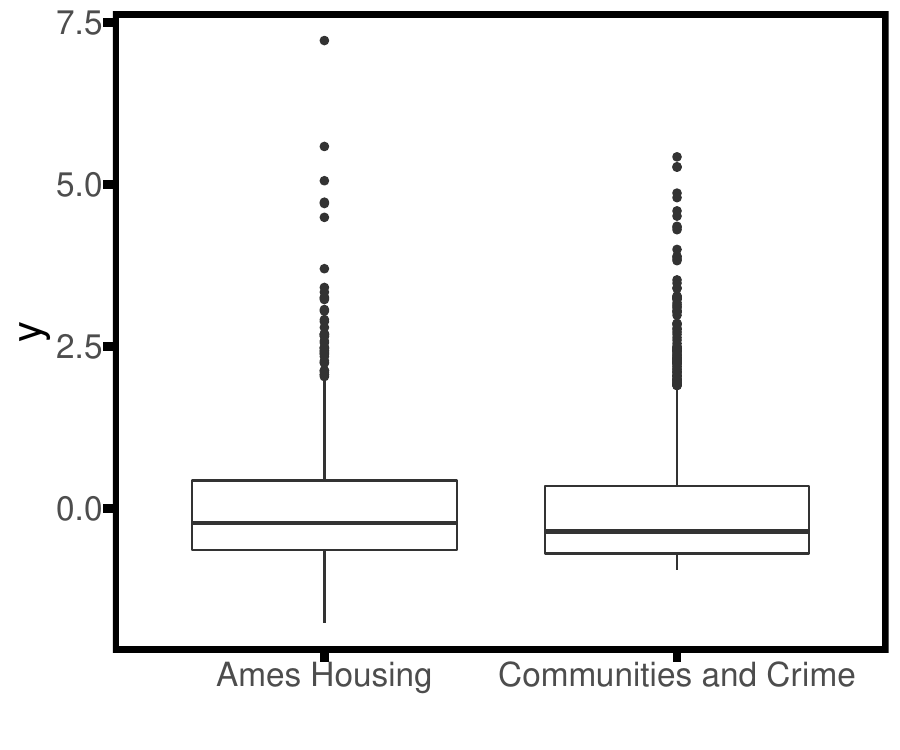}
\caption{The boxplots for the normalized responses in the training data for Ames Housing and Communities and Crime dataset.}
\label{fig:boxplot}
\end{wrapfigure}

\noindent \textbf{Dataset Description:} 
For the Communities and Crime dataset, we analyze the linear relationship between the demographic variables and the community crime number and identify the significant demographic variables. In our analysis, the response variable is the total number of violent crimes per 100K population (ViolentCrimesPerPop). The final dataset includes 99 features and 1993 communities.
For the Ames Housing dataset, we analyze the linear relationship between the sale price of a house and the house's attributes, such as building class, number of rooms, and original and construction year. The dataset includes 1121 observations, where each observation contains a sale price and 37 attributes of the corresponding house. Figure \ref{fig:boxplot} shows boxplots for the normalized responses in the training data for the Ames Housing and Communities and Crime dataset. We can observe that both data have skewed distribution and that the Ames housing data is more spread than the Communities and Crime data. So we may say that both data have heavy-tailed data,

\noindent \textbf{Estimation Results:} 
We randomly select 80\% of data points as the training dataset, and we use the other 20\% as the test dataset. We apply all six methods to the training dataset and obtain the estimated model weights. The parameter settings are the same as in the synthetic data experiment. For evaluation, we calculate the mean square error (MSE) and mean absolute error (MAE) between the predicted responses and the true responses in the test dataset. We also report the number of non-zero estimated model weights as Sparsity. Tables \ref{tab: ameshouse} and \ref{tab: crime} give the MSE, MAE, and Sparsity for all six estimators under different privacy budgets $\epsilon = 0.10$, 0.15, 0.20, 0.25 and 0.30. 

For the Ames Housing dataset, FRAPPE and SgpLAD outperform other square loss based methods since the data has a heavy tail as shown in Figure \ref{fig:boxplot}. FRAPPE and SgpLAD have similar results, however, FRAPPE took a shorter computation time as shown in Figure \ref{fig:time_mse}. More specifically, the FRAPPE estimator reduces MSE by at least about 22\%, and MAE by at least about 31\% compared with square loss based methods.

For the Communities and Crime dataset, FRAPPE outperforms other methods as it is robust and fast. For this dataset, the results are quite sensitive to the initialization point, and hence the MSEs and MAEs for HtPSLR are not stable.

 Hard-threshold methods (DPSLR, HtPSLR, DPIGHT) select few covariates and suffer from a larger MSE. FRAPPE selects a reasonable number of covariates and has the best model performance in terms of both MSE and MAE.
The results show that the proposed FRAPPE has the best performance in estimating as well as in sparsity recovery.    

\begin{table*}
\small
\caption{The testing results for the Ames Housing dataset. We report MSE, MAE, and sparsity under different privacy budgets $\epsilon$ ranging from 0.1 to 0.3. The bold values under the same setting (row) are the two smallest MSEs or MAEs.}
	\label{tab: ameshouse}
	\centering
\begin{tabular}{clcccccc}
  \hline
  \hline
$\epsilon$ & & FRAPPE & SgpLAD & GpLASSO & DPSLR & HtPSLR & DPIGHT \\ 
  \hline
\multirow{3}{*}{0.10} & MSE & \textbf{0.32} & \textbf{0.29} & 1.12 & 32.74 & 1.02 & 0.77 \\ 
 & MAE & \textbf{0.32} & \textbf{0.30} & 0.71 & 3.98 & 0.62 & 0.60 \\ 
 & Sparsity & 33 & 28 & 14 & 4 & 4 & 4 \\ 
  \hline
\multirow{3}{*}{0.15} & MSE & \textbf{0.30} & \textbf{0.29} & 0.87 & 22.72 & 0.69 & 0.58 \\ 
 & MAE & \textbf{0.30} & \textbf{0.30} & 0.62 & 3.15 & 0.52 & 0.52 \\ 
 & Sparsity & 31 & 25 & 22 & 4 & 4 & 4 \\ 
  \hline
\multirow{3}{*}{0.20} & MSE & \textbf{0.28} & \textbf{0.30} & 0.70 & 10.46 & 0.66 & 0.48 \\ 
 & MAE & \textbf{0.29} & \textbf{0.30} & 0.54 & 2.21 & 0.49 & 0.46 \\ 
 & Sparsity & 28 & 23 & 30 & 4 & 4 & 5 \\
  \hline
\multirow{3}{*}{0.25} & MSE & \textbf{0.30} & \textbf{0.29} & 0.57 & 6.59 & 0.78 & 0.44 \\ 
 & MAE & \textbf{0.30} & \textbf{0.30} & 0.48 & 1.79 & 0.51 & 0.44 \\ 
 & Sparsity & 23 & 22 & 34 & 4 & 3 & 5 \\ 
  \hline
\multirow{3}{*}{0.30} & MSE & \textbf{0.32} & \textbf{0.28} & 0.47 & 5.35 & 0.76 & 0.41 \\ 
 & MAE & \textbf{0.32} & \textbf{0.29} & 0.43 & 1.58 & 0.52 & 0.42 \\ 
 & Sparsity & 20 & 23 & 35 & 4 & 4 & 5 \\ 
   \hline
   \hline
\end{tabular}
\end{table*}  

\begin{table*}
\small
\caption{The testing results for the Communities and Crime dataset. We report MSE, MAE, and sparsity under different privacy budgets $\epsilon$ ranging from 0.1 to 0.3. The bold values under the same setting (row) are the two smallest MSEs or MAEs.}
\label{tab: crime}
\centering
\begin{tabular}{clcccccc}
  \hline
  \hline
$\epsilon$ & & FRAPPE & SgpLAD & GpLASSO & DPSLR & HtPSLR & DPIGHT \\ 
  \hline
\multirow{3}{*}{0.10} & MSE & \textbf{0.49} & 1.74 & \textbf{1.08} & 8.83 & 19.27 & 4.85 \\ 
   & MAE & \textbf{0.44} & \textbf{0.54} & 0.71 & 2.65 & 2.16 & 0.78 \\ 
   & Sparsity &  11 & 31 & 8 & 3 & 3 & 3 \\ 
   \hline
\multirow{3}{*}{0.15} &   MSE & \textbf{0.48} & 0.77 & \textbf{0.73} & 1.18 & 36.64 & 4.30 \\  
   & MAE & \textbf{0.44} & \textbf{0.55} & 0.57 & 1.55 & 3.02 & 0.82 \\ 
   & Sparsity &11  & 21  & 60  & 7  & 3 & 4 \\ 
   \hline
\multirow{3}{*}{0.20} &   MSE & \textbf{0.49} & \textbf{0.67} & 0.90 & 1.15 & 47.14 & 4.26 \\
   & MAE & \textbf{0.45} & \textbf{0.52} & 0.63 & 0.75 & 3.46 & 0.81 \\ 
   & Sparsity &10  & 18  & 35  & 3  & 3 & 3 \\ 
   \hline
\multirow{3}{*}{0.25} &   MSE & \textbf{0.51} & 0.76 & \textbf{0.73} & 1.17 & 25.98 & 4.04 \\ 
   & MAE & \textbf{0.46} & \textbf{0.55} & 0.57 & 0.76 & 2.37 & 0.76 \\ 
   & Sparsity & 10  & 0  & 41  & 3  & 3 & 4 \\ 
   \hline
\multirow{3}{*}{0.30} &   MSE & \textbf{0.54} & 0.62 & \textbf{0.50} & 0.99 & 33.25 & 4.29 \\
   & MAE & \textbf{0.47} & 0.50 & \textbf{0.46} & 0.67 & 2.91 & 0.83 \\ 
   & Sparsity & 19 & 30 & 64 & 5 & 3 & 4 \\ 
   \hline
   \hline
\end{tabular}
\end{table*}

\section{Conclusion} \label{sec:conclusion}

In this paper, we studied the robust LAD regression problem with sparsity and privacy constraints. 
We proposed an efficient learning method named FRAPPE, which includes surrogate loss reformulation and three-stage noise injection for fast computation and privacy-preserving.
We showed that with properly selected parameters, our estimators achieve a trade-off between the near-optimal rate of $\sqrt{s\log p/N}$ and $\sqrt{p\log(1/\delta)\log(N\epsilon)}G/(N \epsilon)$.
Extensive numerical studies were conducted and verified that our proposed FRAPPE method outperforms the state-of-the-art privacy-preserving learning algorithms for robust sparse regression analysis.
An interesting extension of our work is to generalize the proposed algorithm to classification problems, such as logistic regression.   

\bibliographystyle{IEEEtran}
\bibliography{reference}


\end{document}